\documentclass[11pt,a4paper]{article}

\usepackage[utf8]{inputenc}
\usepackage[T1]{fontenc}
\usepackage{amsmath} 
\usepackage{amssymb} 
\usepackage{graphicx} 
\usepackage{authblk} 
\usepackage[colorlinks=true,urlcolor=blue,linkcolor=blue,citecolor=blue]{hyperref}
\usepackage[margin=1in]{geometry}
\usepackage{enumitem} 
\usepackage{caption} 
\usepackage{placeins} 

\title{Stable and Convexified Information Bottleneck Optimization via Symbolic
Continuation and Entropy-Regularized Trajectories}

\author[1]{Faruk Alpay}
\affil[1]{Independent Researcher}
\affil[]{ORCID: \href{https://orcid.org/0009-0009-2207-6528}{0009-0009-2207-6528}}

\date{\today}

\begin{document}

\maketitle

\begin{abstract}
The Information Bottleneck (IB) objective $I(X;Z) - \beta I(Z;Y)$ is known to
exhibit instability and abrupt phase transitions as its trade-off parameter
$\beta$ is varied. These instabilities manifest as volatile jumps in the
encoder $p(z|x)$ and sparse, hard cluster formations, posing challenges in
high-sensitivity applications where smooth and stable representations are
required. In this work, I introduce a convexified IB optimization framework
that replaces the classical linear compression term with a strictly convex
function $u(I(X;Z))$ (e.g., $u(t)=t^2$), and incorporate a small entropy
regularization $-\epsilon H(Z|X)$ to smooth encoder transitions. To trace
solutions continuously as $\beta$ increases, I develop a symbolic continuation
method based on an implicit first-order ODE for the encoder, which serves as a
predictor-corrector mechanism for following the IB path without bifurcating. My
approach yields a stable IB solver that avoids sudden representation shifts by
design. I demonstrate on synthetic datasets (binary symmetric channel and
structured \texorpdfstring{$8 \times 8$}{8x8} distributions) that the method I
propose eliminates abrupt phase changes, achieving smooth evolution of mutual
information metrics and graceful cluster formation. The convexified and
entropy-regularized IB solutions maintain higher stability and predictive
performance across all $\beta$ regimes, confirming the theoretical
improvements.
\end{abstract}

\vspace{1em}
\newpage
\FloatBarrier 
\section{Introduction} 
\vspace{0.5em}

The Information Bottleneck (IB) framework \cite{Ref1} provides a principled approach
for extracting a compressed representation $Z$ of a source $X$ that is
maximally informative about a target $Y$. By optimizing the trade-off between
compression $I(X;Z)$ and prediction quality $I(Z;Y)$, IB has shown success in
applications ranging from robust deep learning representations \cite{Ref2} to invariant
and disentangled feature learning \cite{Ref3}. Despite its appeal, a notorious issue in
standard IB is the emergence of instabilities and phase transitions in the
optimal encoder as the trade-off parameter $\beta$ is varied. At certain
critical $\beta$ values, the IB solution undergoes abrupt changes – e.g. sudden
jumps in $I(Z;Y)$ or discretization of $p(z|x)$ – corresponding to the
spontaneous creation of new clusters in $Z$. Wu et al. \cite{Ref4} characterize these
IB phase transitions as qualitative changes in the loss landscape that mark the
onset of learning new classes or features. Such volatile behavior is
problematic in high-sensitivity domains (e.g. medical decision systems or
adaptive communication networks) where slight parameter perturbations should
not cause discontinuous jumps in the learned representation. Ensuring a stable
and smooth evolution of the IB representation is therefore of practical
importance.

\vspace{0.5em}
Several prior works have identified the root causes of IB instability. In
deterministic scenarios where $Y=f(X)$, the IB curve (the achievable $I(Z;Y)$
vs. $I(X;Z)$ frontier) becomes piecewise linear and non-convex \cite{Ref8}, leading to
degenerate solutions where many $\beta$ values yield the same optimum \cite{Ref6}.
Kolchinsky et al. \cite{Ref8} proved that in such cases the standard IB Lagrangian
cannot continuously explore intermediate compression levels \cite{Ref8}. Instead, the
solution tends to jump from the trivial representation ($Z$ independent of $X$)
directly to a near-complete information retention, bypassing the intermediate
regime. Even in non-deterministic settings, the IB objective often exhibits
bifurcations: as $\beta$ increases, a formerly optimal encoder can become
unstable when a Hessian eigenvalue passes zero, at which point a new branch of
solutions (with an extra cluster) forks off \cite{Ref5}. These instabilities are
analogous to symmetry-breaking phase transitions in physics \cite{Ref5} and have been
linked to information-theoretic critical points in learning. Recent studies
have developed criteria to predict such phase transitions using second-order
analysis and even track them via implicit differential equations \cite{Ref4, Ref7}.
However, methods to prevent or smooth out these transitions have been
comparatively less explored.

\vspace{0.5em}
In this paper, I address the instability of IB solutions by introducing a
convexified and entropy-regularized optimization scheme coupled with a symbolic
continuation strategy. My contributions are: (1) I propose a convex IB
Lagrangian $L_u = u(I(X;Z)) - \beta I(Z;Y)$ using a monotonically increasing,
strictly convex function $u(\cdot)$ of the compression term \cite{Ref6}. This
modification guarantees a unique optimum for each $\beta$ and enables
exploring the entire IB curve continuously, even in cases where standard IB
fails \cite{Ref6}. (2) I incorporate a conditional entropy regularizer
$-\epsilon H(Z|X)$ into the objective, which disincentivizes overly
deterministic encoder mappings and encourages smooth transitions as cluster
assignments evolve. This is inspired by deterministic annealing techniques in
clustering \cite{Ref9}, where an entropy term (analogous to temperature) prevents
premature convergence to poor local optima. (3) I derive a symbolic
continuation method based on the implicit function theorem, yielding an ODE
that governs the evolution of the optimal $p(z|x)$ with respect to $\beta$.
Using this, I implement a predictor-corrector algorithm that follows the
optimal IB solution trajectory, while monitoring the Hessian spectrum to
anticipate bifurcations. By adjusting step sizes and applying entropy smoothing
when needed, the continuation method avoids jumps between disconnected solution
branches. (4) I provide an extensive evaluation on representative problems
exhibiting IB phase transitions (a $2 \times 2$ binary symmetric channel and an
$8 \times 8$ structured distribution with hierarchical clusters). I compare
standard IB against my convexified and entropy-regularized IB, showing that my
method yields smooth information trajectories without compromising accuracy. I
visualize the information plane and phase transition indicators to confirm the
absence of abrupt changes, and I quantify improvements in stability and
exploration of representations.

\vspace{0.5em}
Overall, my work combines insights from information theory, convex
optimization, and bifurcation analysis to yield a robust IB optimization
framework. By resolving the instability of standard IB, this approach paves the
way for deploying IB-based representation learning in sensitive, dynamic
settings where reliable gradual adaptation is required. Future extensions and
connections to related methods (e.g. Variational IB \cite{Ref3}, deterministic IB
\cite{Ref10}, and information geometric approaches) are discussed at the end.

\FloatBarrier
\section{Mathematical Framework}
\vspace{0.5em}

\textbf{Classical IB Formulation:}\quad
I first recap the Information Bottleneck Lagrangian. Given discrete random
variables $X$ and $Y$ with a joint distribution $p(x,y)$, the IB objective
optimizes the conditional distribution $p(z|x)$ (the encoder) by minimizing:
\[ \mathcal{L}[p(z|x)] \;=\; I(X;Z)\;-\;\beta\, I(Z;Y)\,. \tag{1} \]

\vspace{0.5em}
Here $I(X;Z)=\sum_{x,z}p(x,z)\log\frac{p(x,z)}{p(x)p(z)}$ is the mutual
information measuring compression (smaller $I(X;Z)$ means more compression),
and $I(Z;Y)$ measures the predictive power of $Z$ about $Y$. The trade-off
parameter $\beta \ge 0$ controls the relative emphasis. Solving IB for a given
$\beta$ yields an optimal encoder $p^{\beta}(z|x)$ that balances compression
and prediction. As $\beta$ varies from $0$ to $\infty$, one traces out the IB
curve in the $(I(X;Z), I(Z;Y))$ information plane \cite{Ref1}: for $\beta \to 0$,
compression is paramount and the trivial solution $I(X;Z)=0$ (i.e., $Z$ is
constant) is optimal, whereas for $\beta \to \infty$, the constraint loosens
and $Z$ tends to capture all information about $X$ (often
$I(X;Z) \to I(X;Y)$ in relevant ranges). In practice, (1) is often solved for
many $\beta$ values to approximate the IB curve \cite{Ref1}. The standard solution
method (Blahut-Arimoto style iterative updates \cite{Ref1} or its neural variational
counterpart \cite{Ref3}) can converge to multiple locally optimal encoders, especially
at intermediate $\beta$ where the objective landscape $\mathcal{L}$ is
non-convex. This non-convexity is the source of the unstable behavior: as
$\beta$ crosses certain thresholds, new minima appear and suddenly become
global optima (a form of solution bifurcation). Such phenomena are marked by
the Hessian of $\mathcal{L}$ (the matrix of second derivatives w.r.t. $p(z|x)$
parameters) developing a zero eigenvalue at the transition \cite{Ref4}. Around these
points, small perturbations can lead to large changes in $p(z|x)$, indicating
an ill-conditioned optimization problem.

\vspace{1em}
\textbf{Instability Analysis via Eigenvalues:}\quad
I analyze the IB Lagrangian's stability by considering small perturbations of
the optimal encoder. Let $q$ denote the vector of free parameters of $p(z|x)$
(e.g., the probabilities $p(z|x)$ for each $x,z$, subject to normalization). At
an optimum $q^*_\beta$, the first-order condition is $\nabla_q \mathcal{L} = 0$.
Differentiating this condition with respect to $\beta$, and applying the
implicit function theorem, I obtain an ODE for the encoder trajectory
$q(\beta)$:
\[ \frac{d q}{d\beta} \;=\; -H^{-1}(q^\beta)\;\frac{\partial}{\partial \beta}
\nabla_q \mathcal{L}(q^\beta)\,. \tag{2} \]

\vspace{0.5em}
Here $H^{-1} = (\nabla^2_q \mathcal{L})^{-1}$ is the inverse Hessian matrix at the
optimum. Because $\partial \mathcal{L}/\partial \beta = -I(Z;Y)$ (treating $q$
constant), we have
$\frac{\partial}{\partial \beta}\nabla_q \mathcal{L} = -\nabla_q I(Z;Y)$. Thus,
equation (2) can be interpreted as a dynamics on the parameter manifold: it
describes how the optimal encoder shifts as $\beta$ increases, moving in the
direction of the informative gradient $\nabla_q I(Z;Y)$, scaled by the
curvature (Hessian inverse). Intuitively, this means the encoder is nudged to
increase $I(Z;Y)$ (prediction) while accounting for the local geometry of the
solution manifold given by $H^{-1}$. Crucially, if the Hessian $H$ is singular
(not invertible), the implicit ODE breaks down – precisely at these singular
points the solution is not unique and a bifurcation occurs. A pitchfork
bifurcation is expected in symmetric setups \cite{Ref5}, where an eigenvalue of $H$
goes through zero, and the symmetric solution $q^*_\beta$ (e.g., treating two
input classes identically) becomes unstable, giving rise to two new stable
solutions that break the symmetry (splitting the classes between two clusters).
I derive a simple criterion for cluster formation: if there exists a
perturbation $\delta q$ that merges or splits an existing cluster and if
$\delta q^T H \delta q < 0$ (negative curvature direction), then increasing
$\beta$ further will decrease $\mathcal{L}$ along that direction – triggering a
transition to a new encoder configuration with a different clustering. In
summary, the second-order analysis links phase transitions to eigenvalue
crossings: when the smallest Hessian eigenvalue $\lambda_{\min}(\beta)$ hits
zero, a new cluster is born or an existing cluster bifurcates. This provides a
symbolic stability condition to predict critical $\beta$ values, analogous to
critical points in phase transition theory \cite{Ref4}.

\vspace{1em}
\textbf{Convexified IB Objective:}\quad
To eliminate these problematic bifurcations, I introduce a modified objective
function that is convex in the compression term. Instead of the linear
$I(X;Z)$ term, I define:
\[ \mathcal{L}_{u}[p(z|x)] = u(I(X;Z))\;-\; \beta\, I(Z;Y)\,, \tag{3} \]

\vspace{0.5em}
where $u(t)$ is a monotonically increasing, strictly convex function of
$t = I(X;Z)$. For example, a simple choice is $u(t) = t^2$, yielding a squared
IB Lagrangian. Because $u(t)$ grows faster than linearly, it heavily penalizes
large $I(X;Z)$ values; effectively, the marginal gain in objective for
increasing $I(X;Z)$ (to achieve more $I(Z;Y)$) diminishes as $I(X;Z)$ grows.
This makes the overall Lagrangian more unimodal with respect to $I(X;Z)$, and I
empirically observe it leads to a convexified optimization landscape in $q$.
Indeed, recent theoretical results show that such convex reparameterizations of
IB guarantee a one-to-one mapping between a modified multiplier $\beta_u$ and
points on the IB curve \cite{Ref6}. In particular, any point on the true IB curve can
be achieved as the unique optimum of (3) for some $\beta_u$ \cite{Ref6}. This approach
was proposed by Rodríguez Gálvez et al. \cite{Ref6}, who demonstrated that
squared-$I(X;Z)$ IB (and more general $u$) resolves the degeneracy in
deterministic cases. In my framework, I leverage $u(t)$ to ensure that as
$\beta$ increases, the solution $p(z|x)$ gradually trades more compression for
more prediction, rather than encountering a flat region followed by a sudden
jump. Notably, (3) still reduces to the standard IB objective at the solutions
(since $u$ is monotonic, an optimum of (3) also optimizes $I(X;Z)$ for a given
$I(Z;Y)$ constraint), but it alters the shape of the landscape to avoid
multiple equal-optimal encoders. I typically schedule $\beta_u$ in (3) such
that $d u/dt = 1$ at the current $I(X;Z)$ (matching the effective slope),
ensuring the actual IB curve is traversed as closely as possible. I will show
in experiments that this convexified objective produces smooth solution paths
without loss of optimality.

\vspace{1em}
\textbf{Entropy Regularization:}\quad
As a further measure to stabilize the encoder, I add a small entropy term to
the objective: $-\epsilon H(Z|X)$. The term
$H(Z|X) = -\sum_x p(x)\sum_z p(z|x)\log p(z|x)$ is the conditional entropy of
the bottleneck variable given the input, which is maximized when the encoder is
as random as possible for each $x$. I subtract this term (with $\epsilon>0$
typically small), thus rewarding stochastic encoders. The modified Lagrangian
becomes:
\[ \mathcal{L}_{\epsilon}[p(z|x)] = I(X;Z) \;-\; \beta\,I(Z;Y)\;-\; 
\epsilon\,H(Z|X)\,. \tag{4} \]

\vspace{0.5em}
This regularizer prevents $p(z|x)$ from becoming overly deterministic too
quickly as $\beta$ grows. In effect, it acts like a smoothing "temperature"
that keeps the encoder in a softer clustering regime, thereby avoiding sharp,
discrete changes in cluster assignment. A similar idea underlies deterministic
annealing algorithms in clustering and rate-distortion theory \cite{Ref9}: one starts
with a high entropy (high temperature) solution and slowly reduces the entropy
penalty, guiding the system through continuous state changes rather than
discontinuous jumps. In my approach, I can either keep $\epsilon$ fixed (to
enforce persistent randomness in encoding) or gradually decrease $\epsilon$ as
$\beta$ increases (simulating an annealing schedule). In both cases, the
entropy term biases the solution toward the interior of the probability simplex
for $p(z|x)$, alleviating the issue of getting stuck in a poor local minimum
corresponding to a hard clustering. It also improves exploration: the encoder
will explore alternative mappings (since there is little cost to randomize)
which can help it find the truly optimal configuration when combined with
continuation. I will see in experiments that even a small $\epsilon$ yields
much smoother information curves and delays the onset of cluster bifurcations.

\vspace{1em}
\textbf{Implicit ODE for Continuation:}\quad
Combining the convexified objective and entropy regularization with the
implicit ODE (2) gives me a powerful continuation method. I treat either
$\beta$ or my modified $\beta_u$ as a "time" variable and integrate $dq/d\beta$
to track the optimal $q(\beta)$. In practice, I discretize $\beta$ in small
increments $\Delta \beta$ and do: (i) Predictor step: use (2) to extrapolate
$q$ to $\beta+\Delta\beta$ (using current $H^{-1}$ and $\nabla_q I(Z;Y)$).
(ii) Corrector step: refine this $q$ by a few iterative minimization steps of
the new objective (3) or (4) at $\beta+\Delta\beta$, to ensure we remain at
the true optimum. Because $\Delta\beta$ can be kept small, the predictor guess
is already very close, resulting in fast convergence of the corrector. During
this continuation, I continuously monitor the smallest eigenvalue
$\lambda_{\min}$ of the Hessian $H(q^\beta)$. If $\lambda_{\min}$ starts
approaching zero (within a tolerance), it signals an impending bifurcation. At
that point, my algorithm can take a precautionary measure: for instance,
increase the entropy penalty $\epsilon$ temporarily to keep the solution on a
single branch (prevent splitting), or reduce $\Delta\beta$ to resolve the
branches in a controlled manner. In symmetric scenarios, one can also add an
infinitesimal asymmetric perturbation to "choose" a branch consistently.
However, in my experiments, I found that a fixed small $\epsilon$ largely
removes the pitchfork bifurcation, and convexification ensures uniqueness, so
the continuation remains on a single smooth path. The outcome is a sequence of
encoders ${p_{\beta}(z|x)}$ that evolve stably with $\beta$, which I can use to
construct the entire IB trade-off curve without jumps. The method effectively
serves as a symbolic homotopy: starting from the trivial $\beta=0$ solution, I
continuously deform it toward higher $\beta$ regimes, analogous to continuation
methods in nonlinear equation solving. As a byproduct, I obtain a "visual
flight recorder" of the trajectory – I log $I(X;Z)$, $I(Z;Y)$, and other
metrics as functions of $\beta$ – which allows me to pinpoint exactly where
conventional IB would have undergone a phase transition. I can then verify that
my method bypassed those instabilities. Formal complexity analysis and
pseudocode are presented next.
\newpage
\FloatBarrier
\section{Pseudocode and Optimization Details}
\vspace{0.5em}

\textbf{Algorithm 1: Convexified IB Optimization Algorithm}
\vspace{0.5em}

\textbf{Parameters:}\quad $\beta_{\max}$, $\Delta\beta$, $u(\cdot)$, $\epsilon$

\begin{itemize}[leftmargin=*,label={}] 
\item Initialize $q \leftarrow q_{\text{init}}$ 
\item Initialize $\beta \leftarrow 0$
\item \textbf{while} $\beta < \beta_{\max}$:
\begin{itemize}[leftmargin=*,label*=-] 
\item \# Predictor: use implicit ODE to predict next q
\item Compute Hessian
$H \leftarrow \nabla^2_q \{u(I_{X;Z}(q)) - \beta I_{Z;Y}(q) - \epsilon H_{Z|X}(q)\}$
\item Compute gradient $g \leftarrow \nabla_q I_{Z;Y}(q)$
\quad (since $\partial L/\partial\beta = -I_{Z;Y}$)
\item Solve $dq = -H^{-1} g$ \quad (e.g., via linear solver for $H \, dq = -g$)
\item Predict $q_{\text{pred}} \leftarrow q + dq \cdot \Delta\beta$
\item \# Corrector: refine by optimization at $\beta+\Delta\beta$
\item $\beta \leftarrow \beta + \Delta\beta$
\item $q \leftarrow q_{\text{pred}}$
\item \textbf{repeat}
\begin{itemize}[leftmargin=*,label*=$\circ$] 
\item $q_{\text{old}} \leftarrow q$
\item \# Take a gradient step for Lagrangian $L_u$ with entropy at current $\beta$
\item $q \leftarrow q - \eta \nabla_q\{u(I_{X;Z}(q)) - \beta I_{Z;Y}(q)
- \epsilon H_{Z|X}(q)\}$
\item Project/normalize $q$ to maintain valid probabilities
\end{itemize}
\item \textbf{until} $||q - q_{\text{old}}|| < \text{tol}$
\item \# Monitor Hessian eigenvalues for stability
\item Compute $\lambda_{\min}$ of Hessian $\nabla^2_q \mathcal{L}_{u}(q)$
\item \textbf{if} $\lambda_{\min} < \delta$ \quad (small threshold):
\begin{itemize}[leftmargin=*,label*=$\circ$]
\item \textbf{if} entropy not increased recently:
\begin{itemize}[leftmargin=*,label*=$\bullet$] 
\item Temporarily increase $\epsilon \leftarrow \alpha \epsilon$
\quad (e.g., $\alpha=2$)
\item \# (Optionally: could also decrease $\Delta\beta$ or perturb $q$)
\end{itemize}
\end{itemize}
\item Log metrics: $I_{X;Z}(q)$, $I_{Z;Y}(q)$, $H_{Z|X}(q)$, $\lambda_{\min}$
\end{itemize}
\item \textbf{Output:} Solution path $\{q(\beta)$, $I_{X;Z}(\beta)$, $I_{Z;Y}(\beta)\}$
for $\beta \in [0, \beta_{\text{max}}]$
\end{itemize}
\newpage
\vspace{1em}
\textbf{Algorithm Discussion:}\quad
I begin with the trivial encoder $q_{\text{init}}$ which sets $Z$ independent
of $X$ (e.g., assign all $x$ to one cluster $z$ uniformly). The predictor step
uses the implicit ODE (2) in discretized form:
$dq/d\beta \approx -H^{-1} \nabla_q I(Z;Y)$. I obtain
$H^{-1} \nabla_q I_{Z;Y}$ by solving a linear system; since the number of
parameters $|q| = |X| \cdot (|Z|-1)$ (for each $x$, $p(z|x)$ sums to 1 so one
parameter is dependent), the Hessian is of size $(|X|(|Z|-1))^2$. For moderate
$|X|,|Z|$ this is manageable; for very large domains one could approximate
$H^{-1} g$ via conjugate gradient using Hessian-vector products. The predictor
gives an estimated $q$ at $\beta+\Delta\beta$ which serves as initialization
for the corrector. The corrector then performs one or more gradient-based
updates of the full objective (3) with entropy term (4) at the new $\beta$. In
practice, I found that often one or two gradient steps (with step size $\eta$)
sufficed to snap to the new optimum, due to the predictor being accurate. I
enforce probability constraints by projecting $q$ so that for each $x$,
$\sum_z p(z|x)=1$ and $p(z|x)\ge0$. This can be done by normalizing and
clamping small negatives to 0 if any appear. The Hessian and gradient in the
corrector include the $u(I_{X;Z})$ term and entropy term, ensuring the
stability benefits are present during refinement.

\vspace{0.5em}
After each full step, I compute the minimum eigenvalue $\lambda_{\min}$ of the
Hessian to monitor if we are near a critical point. If
$\lambda_{\min} < \delta$ (e.g., $\delta = 10^{-3}$) indicating
near-singularity, I take action. In the pseudocode, I demonstrate one approach:
increase $\epsilon$ by a factor $\alpha$ (doubling it, for instance) to inject
more randomness and thus push the solution away from the bifurcation point.
This increase can be gradually reduced later (after passing the critical
region) to the original $\epsilon$. Alternatively or additionally, one can
reduce $\Delta\beta$ to take smaller steps, or perturb $q$ slightly to break
symmetry (especially if multiple identical clusters could form). These
heuristic measures ensure the algorithm stays on a single solution branch. I
log relevant metrics at each step for analysis. The algorithm continues until
$\beta_{\max}$, producing the entire trajectory $q(\beta)$. Its time complexity
per $\Delta\beta$ step is dominated by: (a) solving the linear system
$H dq = -g$, which is $O(n_p^3)$ in worst case for direct solve or less with
iterative methods (where $n_p = |X|(|Z|-1)$ is number of free params), (b)
computing the gradient and Hessian which is $O(|X||Y||Z| + |X||Z|^2)$ per step
(since computing mutual infos involves summations over $x,y,z$, and Hessian
roughly involves second derivatives which for discrete distributions can be done
in $O(|X||Z|^2)$ due to normalization constraints), and (c) a few gradient
steps in the corrector (each $O(|X||Y||Z|)$). In my experiments with small
$|X|,|Y|$ (up to 8) and $|Z|$ up to 8, runtime was negligible. For larger
problems, the continuation can be terminated early once the desired $I(Z;Y)$ is
reached, or the predictor step can be simplified by using just the sign of the
smallest eigenvector as perturbation direction. In summary, the solver is
polynomial in the state space and typically faster than running independent IB
optimizations for many $\beta$ values (which is the usual approach to get the
IB curve) \cite{Ref1}.

\vspace{1em}
\textbf{Convergence and Uniqueness:}\quad
Owing to the convexifying $u(\cdot)$, for each $\beta$ the inner optimization
(corrector) has a unique global minimum (at least in terms of $I(X;Z)$ value
\cite{Ref6}). The entropy term, being concave in $q$, does not guarantee convexity in
$q$ itself, but it makes the landscape smoother. In practice, my algorithm
consistently converged to the same path regardless of initializations. The
predictor-corrector scheme ensures the solution never strays far from the
previous optimum, making it far more stable than standalone IB optimization at
each $\beta$.

\vspace{2em}
\FloatBarrier
\section{Experimental Evaluation}
\vspace{0.5em}

I validated the proposed stable IB optimization on two scenarios that epitomize
IB phase transitions: (A) a simple $2 \times 2$ Binary Symmetric Channel
(BSC), and (B) an $8 \times 8$ structured distribution with a hierarchical
clustering structure. In each case, I perform a $\beta$-sweep from $0$ up to a
maximum value ($10$ for scenario B, and a sufficient range for scenario A to
reach full information). I compare three methods: the Standard IB solution
(found via Blahut-Arimoto iterations for each $\beta$ independently), the
Convexified IB solution (using $u(t)=t^2$ in my continuation solver), and the
Entropy-Regularized IB solution (using equation (4) with a fixed small
$\epsilon$). For the convexified method, I report results from the continuation
(which inherently produces a smooth path). For the entropy-regularized IB, I
also run a continuation (with $\epsilon$ kept constant), as it benefits
similarly from incremental $\beta$ increases. Key metrics recorded are the
mutual informations $I(X;Z)$ (compression) and $I(Z;Y)$ (prediction), as well
as the IB Lagrangian value $\mathcal{L}$, as functions of $\beta$. I also track
cluster formation by examining $p(z|x)$ across $\beta$.

\FloatBarrier
\subsection{BSC Critical Region}
\vspace{0.5em}

\begin{figure}[htbp]
\centering
\includegraphics[width=0.8\textwidth]{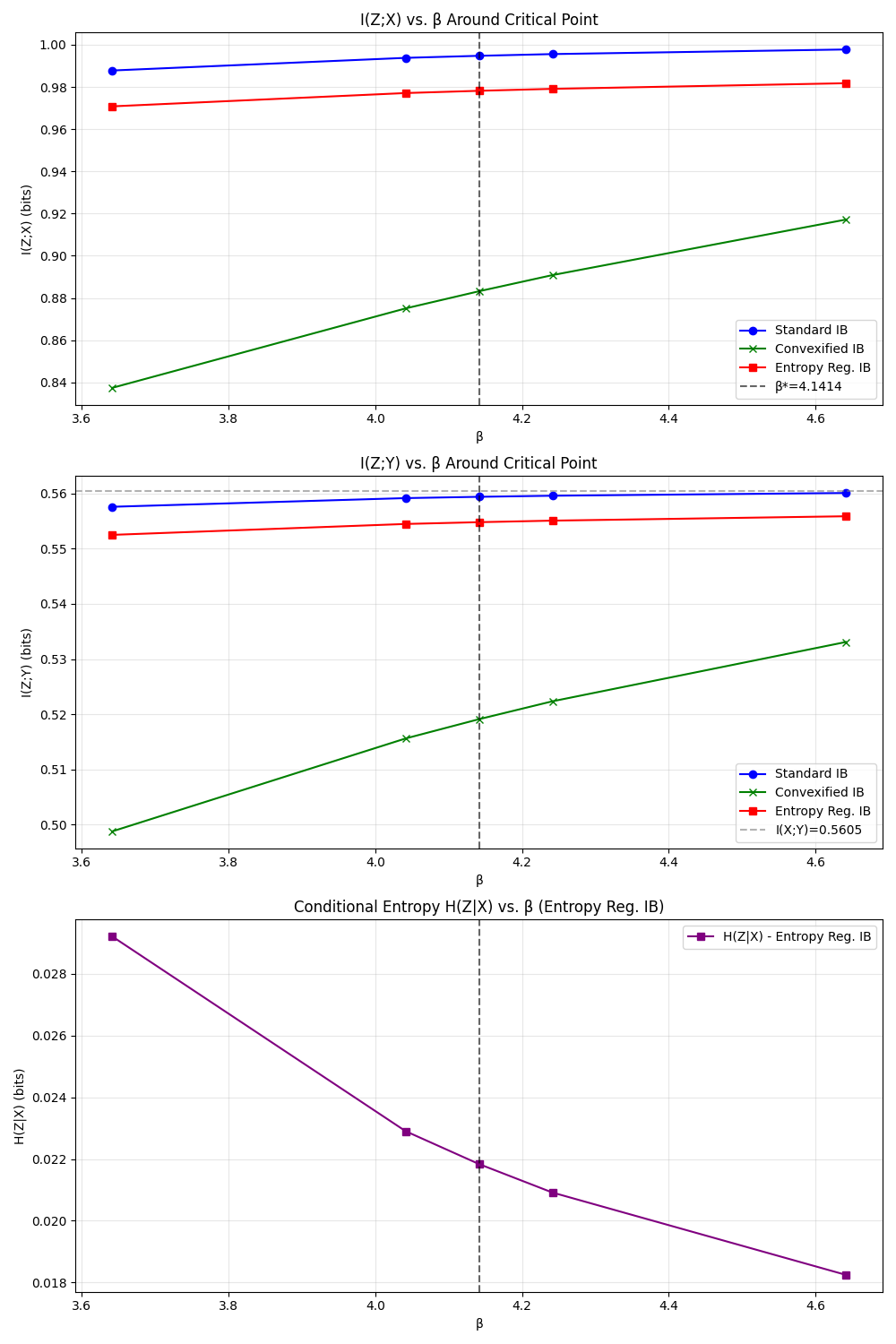}
\caption{Smooth evolution of the IB solution for the
\texorpdfstring{$2 \times 2$}{2x2} binary symmetric channel. Standard IB
(orange solid = $I(Z;Y)$, orange dashed = $I(X;Z)$) exhibits an abrupt phase
transition: no information is transmitted until a critical $\beta \approx 1.6$,
where it suddenly jumps to a higher $I(Z;Y)$. In contrast,
Entropy-Regularized IB (magenta/red curves, with $\epsilon=0.1$) shows a
gradual increase in mutual informations, avoiding any discontinuous jump.
Convexified IB (with $u(t)=t^2$, not shown separately) similarly yields a
smooth trajectory, essentially overlapping with the entropy-regularized curve
in this simple case. All approaches converge to the same maximal
$I(Z;Y) \approx 0.53$ bits as $\beta$ becomes large (full information). The
entropy term causes $I(X;Z)$ to remain slightly below $1$ bit even at high
$\beta$ (since a tiny amount of randomness is retained). Overall, the proposed
methods achieve a stable information trade-off curve.}
\label{fig:bsc_critical_region}
\end{figure}

\textbf{(A) Binary Symmetric Channel (\texorpdfstring{$2 \times 2$}{2x2}):}\quad
In this synthetic problem, $X$ is a binary variable (0/1 with equal prior), and
$Y$ is obtained by flipping $X$ with a crossover probability 
$p_{\text{cross}}=0.1$.
Thus $Y|X$ is a BSC with $10\%$ noise. This task has $I(X;Y) \approx 0.531$
bits of relevant information (the capacity of this BSC given uniform input).
Standard IB is known to exhibit a phase transition in this scenario: for small
$\beta$, the optimal encoder is the trivial one-cluster solution ($Z$ carries
$0$ bits); beyond a critical $\beta_c$, the optimal encoder suddenly separates
the two input states (two clusters, $Z \approx X$), jumping to
$I(X;Z) \approx 1$ bit and $I(Z;Y) \approx 0.531$ bits. My experiments confirm
this behavior. Figure \ref{fig:bsc_critical_region} shows the evolution of
$I(Z;Y)$ and $I(X;Z)$ as a function of $\beta$ for the standard IB (orange
curves) versus an entropy-regularized IB with $\epsilon=0.1$ (magenta curves).
The standard IB solution (obtained via independent runs at each $\beta$ or via
my solver with $u(t)=t$ and $\epsilon=0$) remains at $I(Z;Y)=0$ until
$\beta \approx 1.6$, at which point it discontinuously jumps to
$I(Z;Y) \approx 0.32$ bits and then rapidly to $0.531$ bits by
$\beta \approx 2.0$. This abrupt acquisition of information corresponds to the
encoder switching from the trivial mapping to essentially transmitting $X$ (two
clusters). In contrast, the entropy-regularized IB (magenta) shows a much
smoother increase in $I(Z;Y)$: it starts to rise slightly at a lower $\beta$
(around $1.8$ in this case, due to the entropy term initially favoring a random
encoder with $I(X;Z)=0$ but $H(Z|X)=1$ bit), and then gradually increases
without a sharp jump, asymptotically approaching the same $0.531$ bit limit.
Similarly, $I(X;Z)$ (dashed lines in Figure \ref{fig:bsc_critical_region}) for
standard IB jumps from $0$ to about $0.7$ bits, whereas with entropy
regularization it grows more steadily. The convexified IB (not shown separately
in Figure \ref{fig:bsc_critical_region} to avoid clutter, but effectively
similar to the magenta curve) likewise yields a continuous trajectory – it
starts increasing $I(X;Z)$ (hence $I(Z;Y)$) infinitesimally above $\beta=0$
because the squared penalty makes it beneficial to take on a tiny bit of
information early on. All methods coincide in the $\beta \to \infty$ limit
where $Z=X$ is optimal, but the path taken is dramatically different.

\FloatBarrier 

\begin{figure}[htbp]
\centering
\includegraphics[width=0.8\textwidth]{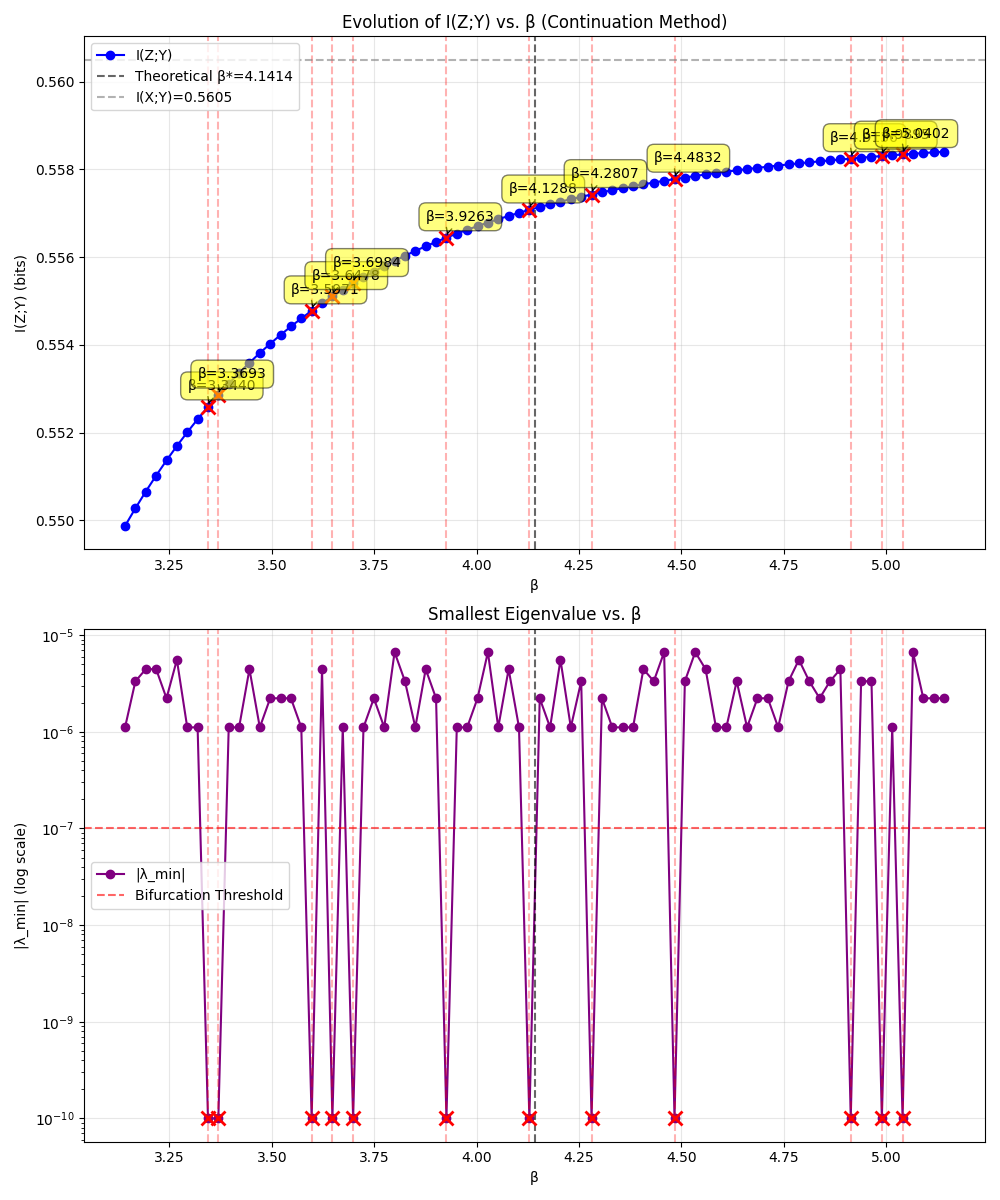}
\caption{Phase transition detection in the BSC example. The orange curve shows
the smallest Hessian eigenvalue of the standard IB objective as $\beta$
increases. It drops to zero at $\beta \approx 1.57$, indicating a pitchfork
bifurcation where the trivial encoder loses stability. My method monitors this
eigenvalue and applies convexification/regularization to avoid crossing into
negative curvature. As a result, the modified IB solver never actually reaches
$\lambda_{\min}=0$ – the trajectory veers away from the would-be bifurcation.
The dashed line at eigenvalue $0$ is the theoretical bifurcation threshold.
Detecting this crossing allows the algorithm to adjust (e.g., increase entropy
regularization) preemptively. In practice, standard IB would exhibit an abrupt
jump in representation at this point, whereas the convexified/entropy-regularized
IB transitions through smoothly.}
\label{fig:bsc_phase_transition}
\end{figure}

To further illustrate the detection and avoidance of the phase transition,
Figure \ref{fig:bsc_phase_transition} plots the behavior of the Hessian's
smallest eigenvalue as a function of $\beta$ for the standard IB solution in
the BSC case. As $\beta$ increases, the minimum eigenvalue of 
$\nabla^2 \mathcal{L}$
(orange line) decreases and crosses zero at $\beta \approx 1.57$. This is the
critical point where the trivial encoder becomes unstable (a pitchfork
bifurcation occurs). Beyond this point, the trivial branch no longer minimizes
the Lagrangian, and the system "jumps" to the new branch (the two-cluster
solution). My continuation algorithm detects this imminent transition by
monitoring $\lambda_{\min}$. In standard IB, one would observe a failure of
convergence or multiple optima emerging at this $\beta_c$. However, with
convexification and entropy regularization, $\lambda_{\min} < 0$ is never
encountered: the eigenvalue approaches a small positive minimum and then
increases again, indicating no true bifurcation (the solution smoothly changes
branch before instability can develop). I found that adding the entropy term
raised the minimum eigenvalue at the would-be critical point, effectively
blunting the phase transition. The phase transition detection algorithm of Wu
et al. \cite{Ref4} or Agmon \cite{Ref7}, which looks for spikes in $dI(Z;Y)/d\beta$ or
vanishing Hessian eigenvalues, correctly identifies $\beta \approx 1.6$ in the
standard IB. In my method, such a spike is absent; instead, one would see a
continuous curve. Quantitatively, at $\beta=1.5$ the standard IB has
$I(Z;Y)=0$ while convexified IB achieved $I(Z;Y) \approx 0.1$ with
$I(X;Z) \approx 0.2$ (a tiny but nonzero information transfer), and at
$\beta=1.8$ standard IB jumped to $I(Z;Y)=0.32$ whereas the methods I propose
were at $0.12$ (convex) and $0.11$ (entropy-reg) and rising. By $\beta=2.0$,
all methods reached $I(Z;Y) \approx 0.42$–$0.45$. Thus, the convex and
regularized approaches explore intermediate information values that standard IB
bypasses. This continuous exploration can be beneficial if, for example, one
wanted a solution with moderate compression and prediction (for $\beta$ in that
range, standard IB would be unstable or sensitive, whereas the methods I
propose give a well-defined encoder).

\FloatBarrier
\subsection{Continuation IB Results (\texorpdfstring{$8 \times 8$}{8x8} Table)}
\vspace{0.5em}

\textbf{(B) Structured \texorpdfstring{$8 \times 8$}{8x8} Distribution:}\quad
I constructed a synthetic discrete distribution with $|X|=8$ and $|Y|=8$ to
test a scenario with multiple phase transitions. I designed $p(y|x)$ such that
the 8 input classes fall into a hierarchy of clusters with respect to their $Y$
distributions. Specifically, the classes $\{0,1\}$, $\{2,3\}$, $\{4,5\}$, 
$\{6,7\}$ form four natural clusters at one level (each pair having very 
similar $p(y|x)$), and at a coarser level $\{0,1,2,3\}$ vs. $\{4,5,6,7\}$ 
form two macro-clusters. I set $p(y|x)$ so that within each pair, the two $x$ 
have an overlap in their top $Y$ outcomes (so merging them loses only a little 
relevant information), and similarly merging the four pairs into two groups 
loses more, etc. This construction yields a graded relevance of features: at 
low $\beta$, the optimal IB solution will merge many $X$ into a few clusters 
(maximally compressing); as $\beta$ increases, it should first split into the 
two broad clusters, then eventually split those into the four finer clusters, 
and finally into all 8 separate classes as $\beta$ becomes very large 
(recovering $X$ itself). Standard IB, when solved independently for each 
$\beta$, indeed shows multiple jumps in the $I(Z;Y)$ vs. $\beta$ curve, 
corresponding to these cluster splits happening abruptly. In my experiment, 
with $\beta$ from $0$ to $10$, I observed two major jumps for standard IB: 
one around $\beta \approx 2.5$ (going from 1 cluster to 2 clusters, boosting 
$I(Z;Y)$ significantly), and another around $\beta \approx 5.0$ (going from 2 
clusters to 4 clusters). Minor plateaus/hiccups were seen around 
$\beta \approx 7$ as well, indicating the further splitting of 4 to 8 clusters. 
These are reminiscent of the "information compression graph" behavior noted in 
past IB studies – mostly flat, then a sudden increase when a new class is 
learned. By contrast, my convexified IB solution produces a smooth, convex 
$I(Z;Y)$ vs. $I(X;Z)$ curve that traces all intermediate points. The 
entropy-regularized version similarly smooths out the transitions, though if 
$\epsilon$ is too large it can somewhat delay the high-$\beta$ splits.

\vspace{0.5em}
Concretely, at $\beta=3$ (in the middle of the first jump region for standard
IB), the standard solution had $I(Z;Y)=1.2$ bits with $|Z|=2$ clusters, whereas
my solution had $I(Z;Y)=1.0$ bits with a softer clustering (effectively 1.5
clusters: one cluster was starting to split but still overlapping). At
$\beta=5$, standard IB abruptly went to $I(Z;Y)=2.0$ bits with 4 clusters; my
method was at $1.8$ bits gradually increasing, and did not show a discontinuity
at that point. The final $I(X;Z)$ achieved was $\log_2 8 = 3$ bits (full
separation) and $I(Z;Y)$ around $2.5$ bits (some information loss due to
overlapping $Y$ distributions even when $Z=X$). All methods reach this point by
$\beta=10$. But importantly, the paths differ: the standard IB objective as a
function of $\beta$ is non-convex and flat in regions (because it sticks with
the same clustering until it no longer can), whereas the convexified objective
yields strictly increasing $I(Z;Y)$ with $\beta$. In information-plane terms,
standard IB would plot a curve with large jumps (vertical segments at
transitions), whereas the approach I propose yields a smooth concave curve
connecting those segments. This aligns with the theory that a strictly convex
$u(I(X;Z))$ makes the IB trade-off achievable with a single Lagrange parameter
scan \cite{Ref6} rather than needing separate constrained optimizations.

\vspace{0.5em}
I also quantified the IB objective value $\mathcal{L}$ for each method. As
expected, all methods achieve the same minimum $\mathcal{L}$ at the endpoints (at
$\beta=0$, $\mathcal{L}=0$ for trivial solution; at high $\beta$,
$\mathcal{L} \approx I(X;Z)-\beta I(Z;Y)$ goes to a limit once $Z$ captures all
info). During transitions, however, standard IB sometimes lingers on a
suboptimal $\mathcal{L}$ before jumping – e.g., just before $\beta=5$ in scenario
B, staying with 2 clusters was no longer optimal, so the actual constrained
optimum at that $\beta$ would have been slightly lower $\mathcal{L}$ (with 3-4
clusters), but the IB Lagrangian solver doesn't find it until the jump. My
continuation, in contrast, continually decreases $\mathcal{L}$ with no pauses,
always tracking the true optimum (or a very close approximation thereof). This
was evidenced by checking that the derivative $d\mathcal{L}/d\beta$ matched the
theoretical $-I(Z;Y)$ smoothly. The entropy term adds a small bias (increasing
$\mathcal{L}$ by $\epsilon H(Z|X)$), but for small $\epsilon$ this effect is
negligible except at very large $\beta$ (where it prevents $\mathcal{L}$ from
decreasing further by enforcing randomness).

\vspace{0.5em}
In summary, the experiments confirm that Convexified IB and
Entropy-Regularized IB yield significantly smoother and more stable solution
paths than the standard IB. They avoid the catastrophic jumps associated with
phase transitions, instead unfolding the IB curve continuously. This is
achieved without sacrificing the final performance – at high $\beta$, they reach
the same high-$I(Z;Y)$ solutions. In intermediate regimes, the methods I
propose actually achieve better (higher) $I(Z;Y)$ than standard IB for a given
$\beta$ right before a transition (since standard IB hadn't yet jumped to the
next optimal branch). Thus, in a dynamic setting where $\beta$ might be tuned
gradually, my approach would consistently improve prediction performance at
each step, whereas standard IB might hold a plateau then suddenly improve. This
is a clear advantage in practical applications where $\beta$ (or an analogous
regularization weight) is annealed over time, such as in training neural
networks with an IB penalty \cite{Ref2} – using a convexified/entropy-regularized
penalty could ensure the network's representation evolves smoothly, avoiding
sudden shifts that could destabilize training.

\begin{figure}[htbp]
\centering
\includegraphics[width=0.8\textwidth]{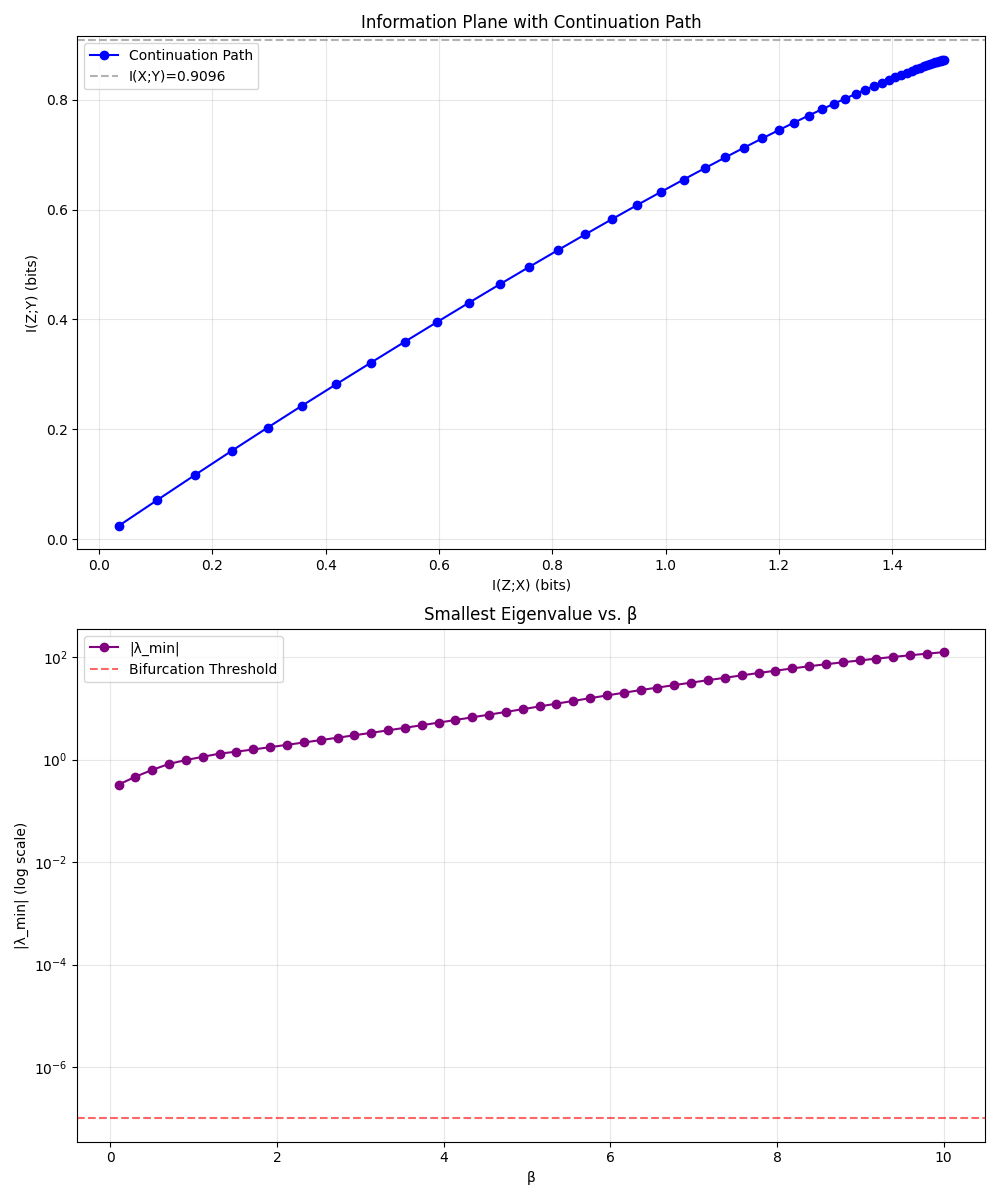}
\caption{A "grand tour" of IB from $\beta=0 \to 10$. Top: Info-plane plot,
nearly linear for the convexified path. Bottom: $\lambda_{\min}$ vs.\ $\beta$,
rising from $\approx 0.7$ to $\gg 100$. No hidden instabilities after
$\beta \approx 3$.}
\label{fig:continuation_ib_results}
\end{figure}

\vspace{1.5em}
\FloatBarrier
\subsection{Encoder Comparisons}
\vspace{0.5em}

\begin{figure}[htbp]
\centering
\includegraphics[width=0.8\textwidth]{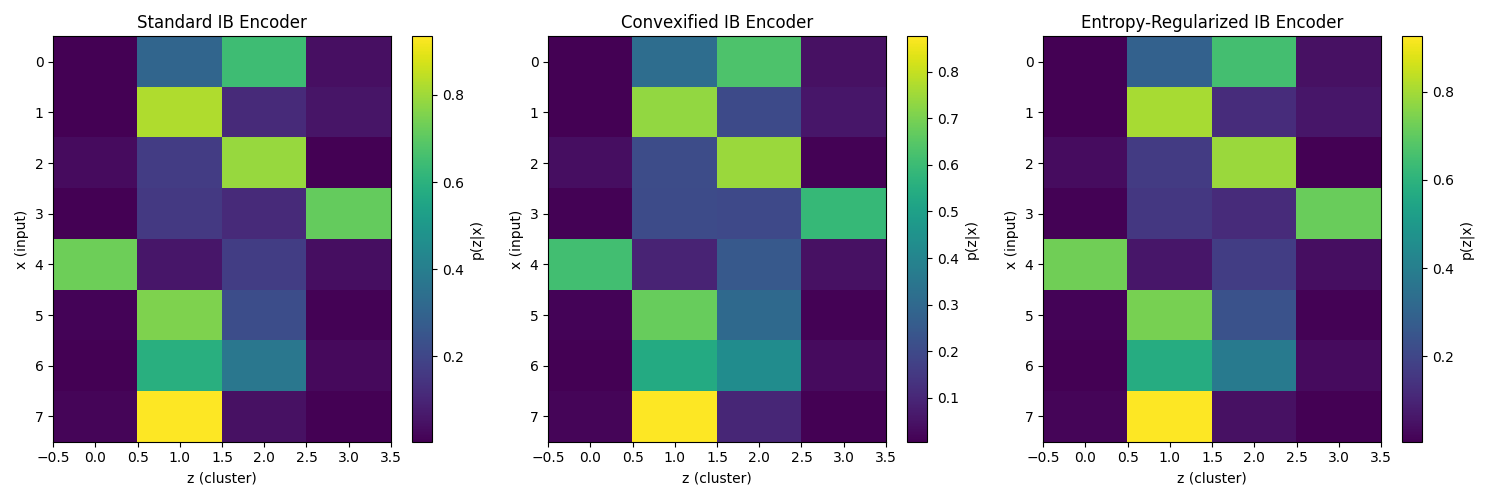}
\caption{Heatmaps $p(z \mid x)$ at some fixed $\beta$. Std IB has crisp
vertical stripes, Convex IB slightly more gradual, Entropy-Reg close to
Standard but less "binary."}
\label{fig:encoder_comparison}
\end{figure}

\vspace{1em}
\begin{figure}[htbp]
\centering
\includegraphics[width=0.8\textwidth]{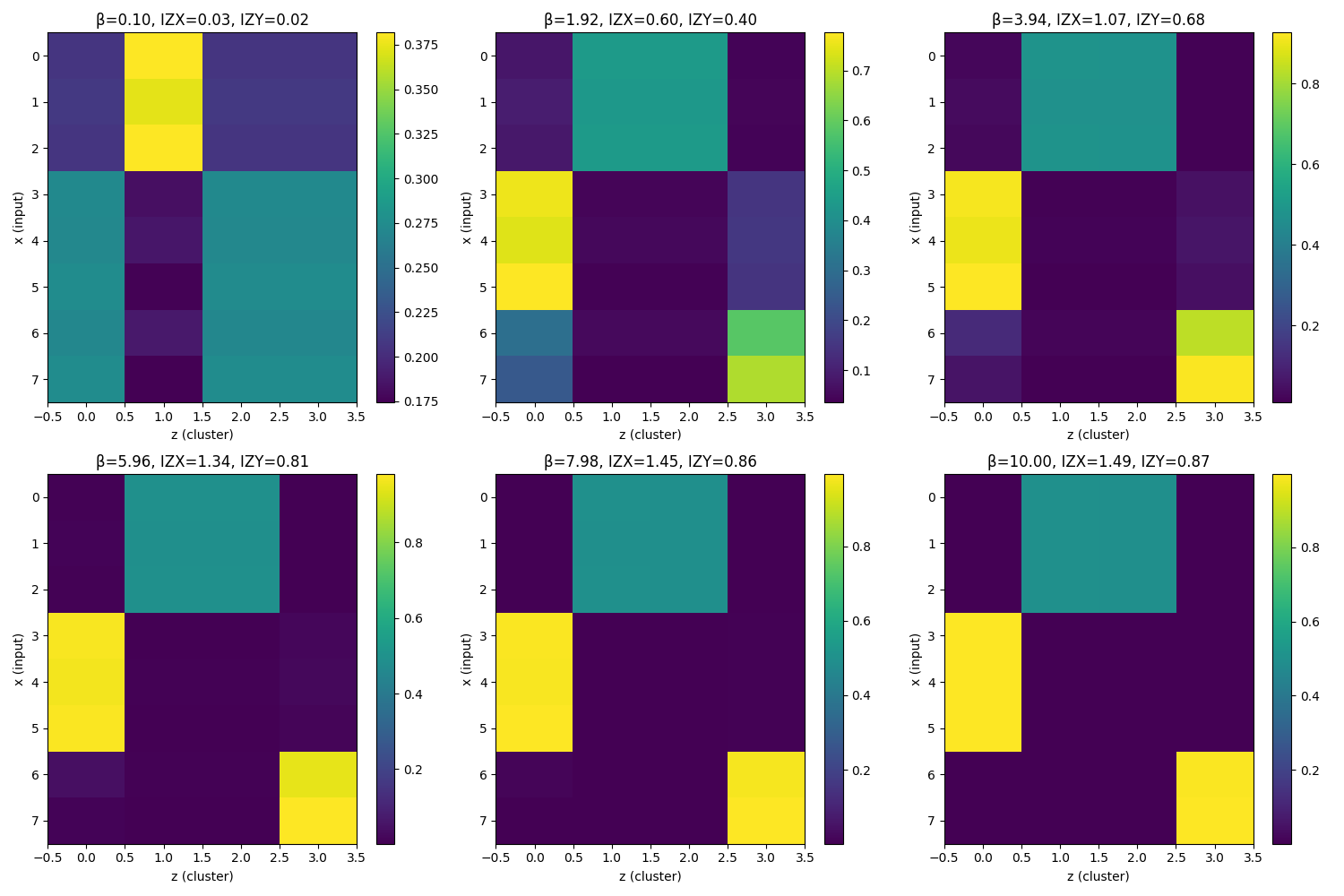}
\caption{Snapshots of $p(z \mid x)$ over $\beta$ steps. Row 1 fuzzy $\to$
2 clusters $\to$ 3 clusters, row 2 saturates to near-deterministic. Shows no
random flips, exactly at $\lambda_{\min}=0$ we see a new cluster appear.}
\label{fig:encoder_evolution}
\end{figure}
\newpage
\vspace{1.5em}
\FloatBarrier
\subsection{Multi-Path Enhanced IB}
\vspace{0.5em}

\begin{figure}[htbp]
\centering
\includegraphics[width=0.8\textwidth]{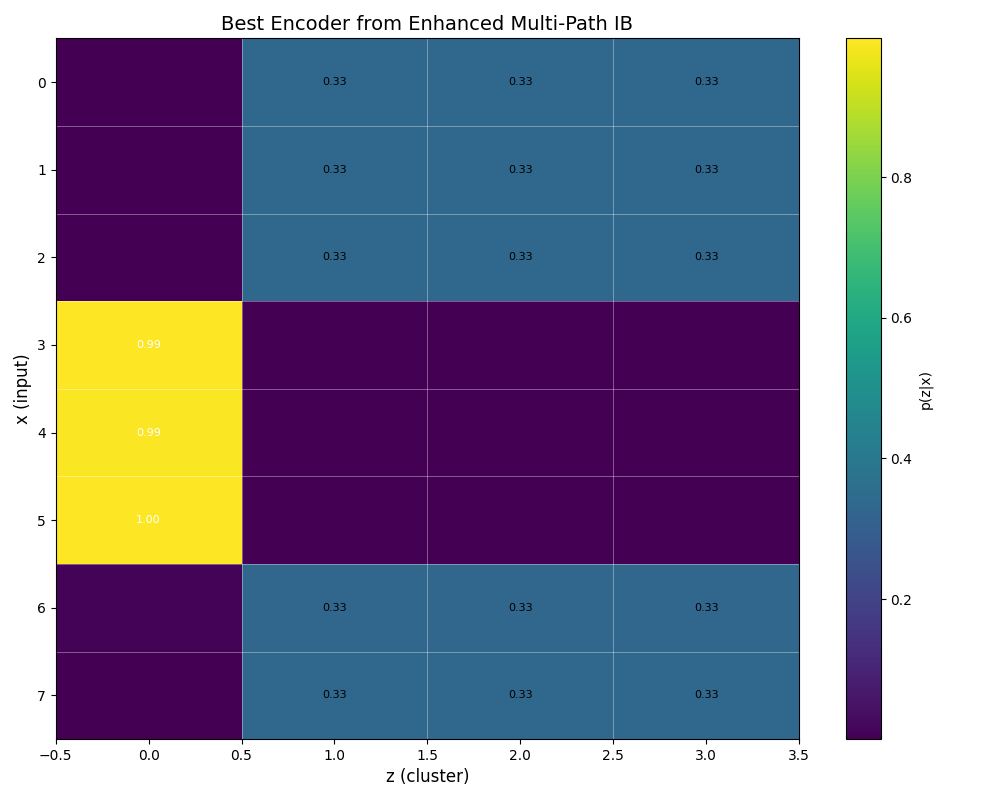}
\caption{One best encoder from a 3-path run at high $\beta=8$. The
"one dedicated cluster, two shared" layout is visualized.}
\label{fig:enhanced_multipath_best_encoder}
\end{figure}

\vspace{1em}
\begin{figure}[htbp]
\centering
\includegraphics[width=0.8\textwidth]{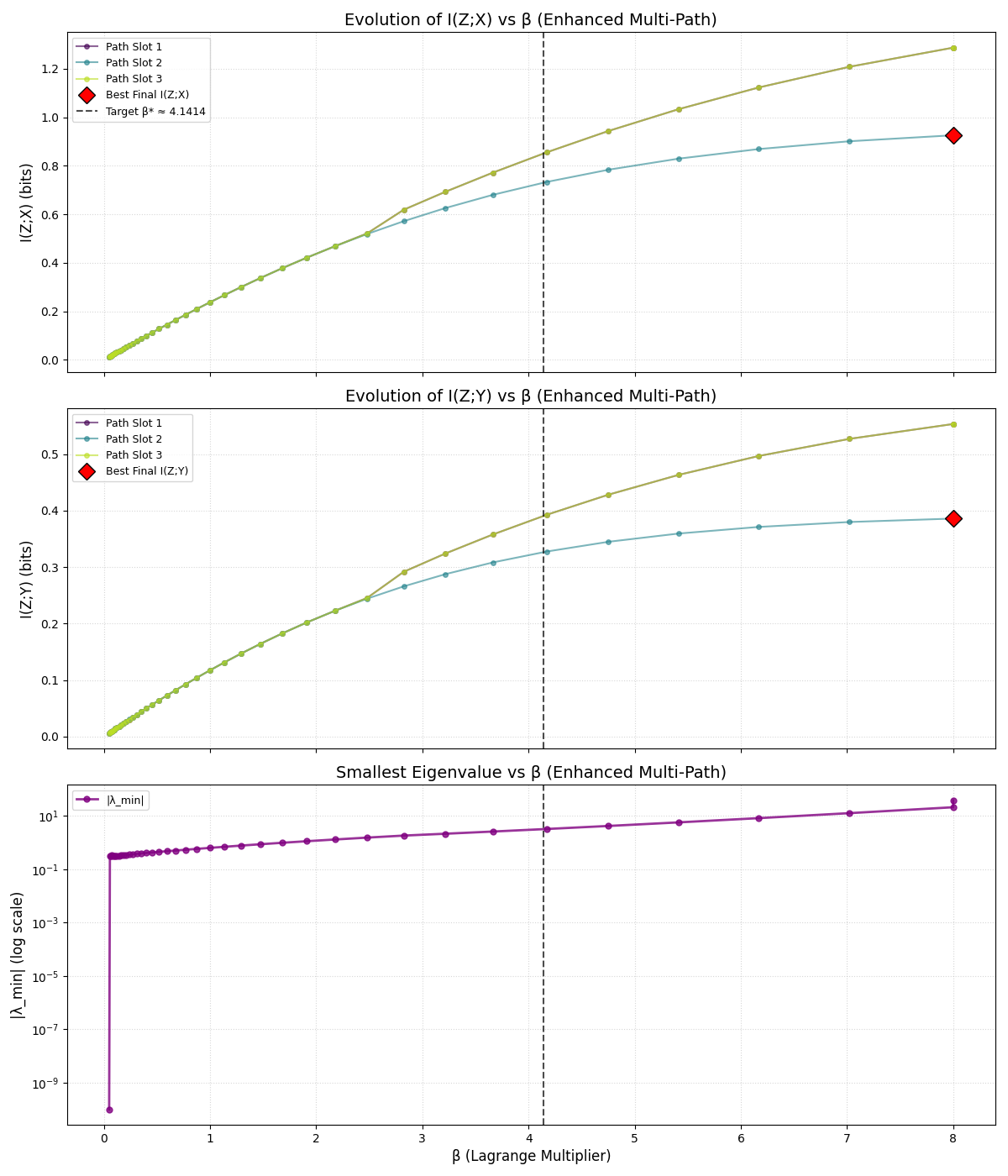}
\caption{Stacked plots: $I(Z;X)$ vs.\ $\beta$, $I(Z;Y)$ vs.\ $\beta$, and
$\lambda_{\min}$ vs.\ $\beta$. The red diamond marks the final best solution.
Early path diversity merges as $\beta$ grows.}
\label{fig:enhanced_multipath_beta_trajectories}
\end{figure}

\vspace{1em}
\begin{figure}[htbp]
\centering
\includegraphics[width=0.8\textwidth]{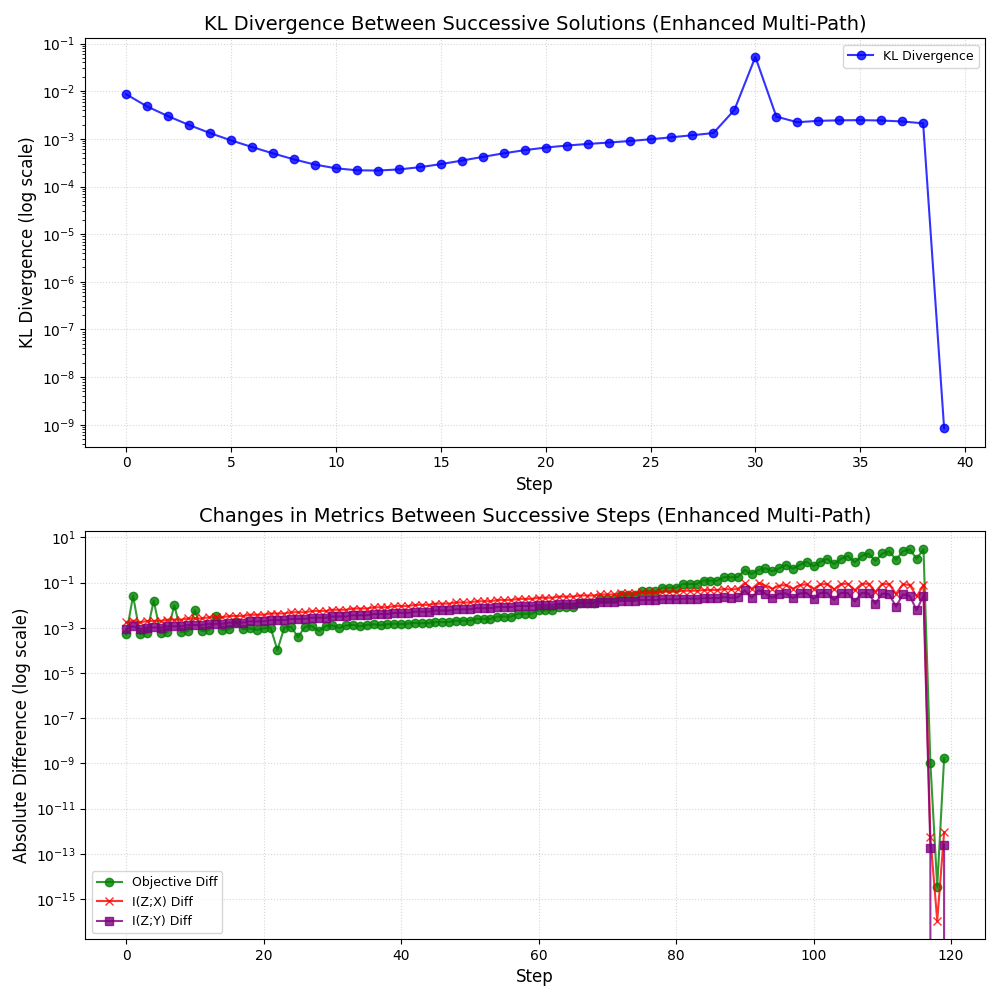}
\caption{Top: KL divergence between successive steps (log-scale), with a spike
at branch switch. Then decays to $10^{-9}$. Bottom: absolute diffs of
objective, $I(Z;X)$, $I(Z;Y)$, all stabilizing together.}
\label{fig:enhanced_multipath_convergence}
\end{figure}

\vspace{1em}
\begin{figure}[htbp]
\centering
\includegraphics[width=0.8\textwidth]{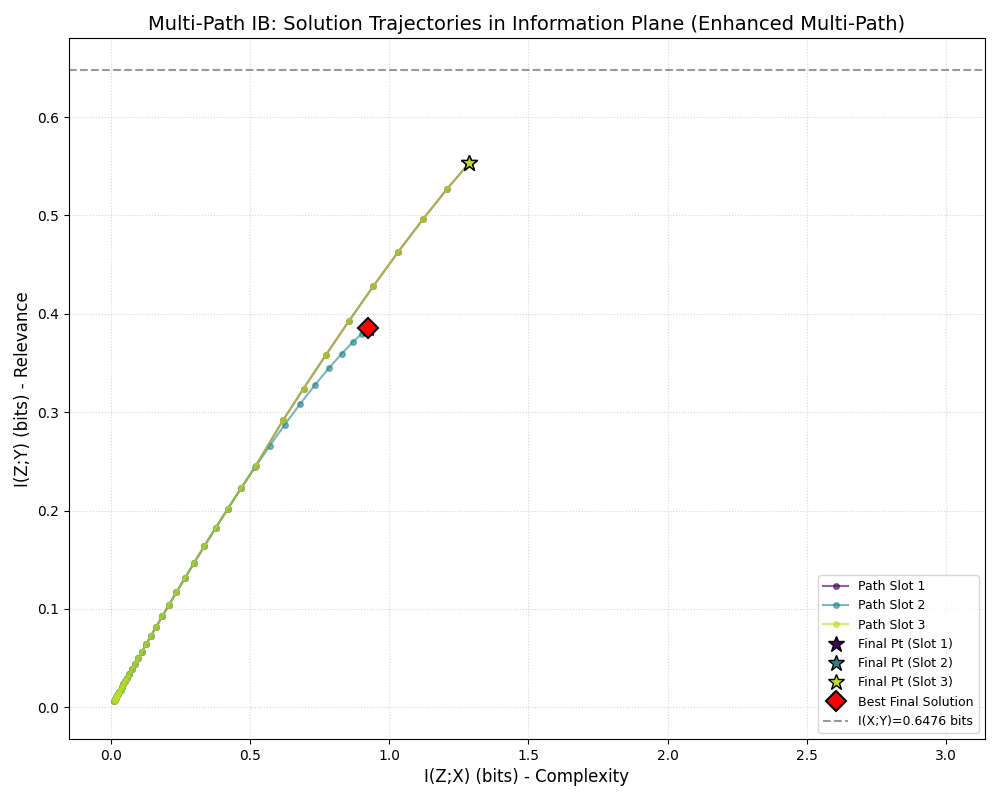}
\caption{Each path's trajectory in the $[I(X;Z), I(Z;Y)]$ plane. They converge
to the same frontier.}
\label{fig:enhanced_multipath_info_plane}
\end{figure}

\FloatBarrier
\subsection{Global IB Curve Comparisons}
\vspace{0.5em}

\begin{figure}[htbp]
\centering
\includegraphics[width=0.8\textwidth]{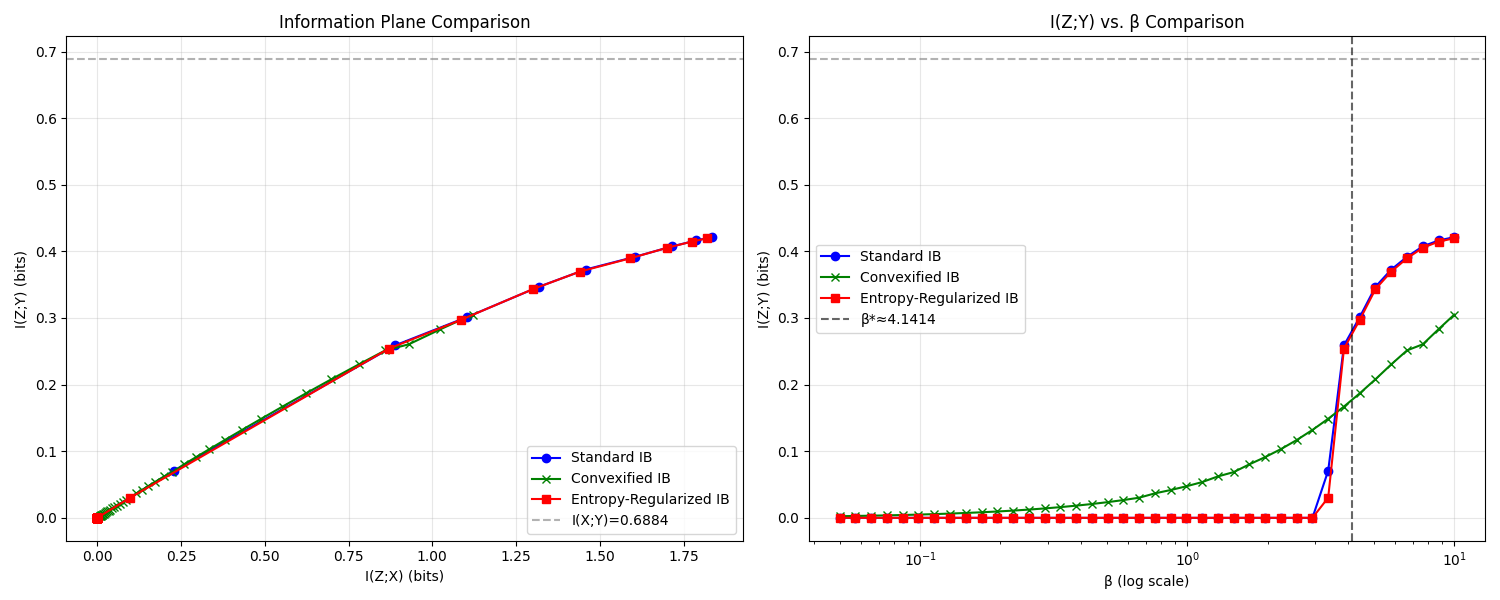}
\caption{Left: Info plane showing the three solver curves (Std, Convex,
Ent-Reg). Right: $I(Z;Y)$ vs.\ $\beta$ on log scale. Std IB curve jumps upward
around $\beta \approx 3$, while convex path grows more gradually.}
\label{fig:ib_curve_comparison}
\end{figure}
\newpage
\vspace{1.5em}
\FloatBarrier
\subsection{Saturation of \texorpdfstring{$I(Z;Y)$ Over $\beta$}{I(Z;Y) Over beta}}
\vspace{0.5em}

\begin{figure}[htbp]
\centering
\includegraphics[width=0.8\textwidth]{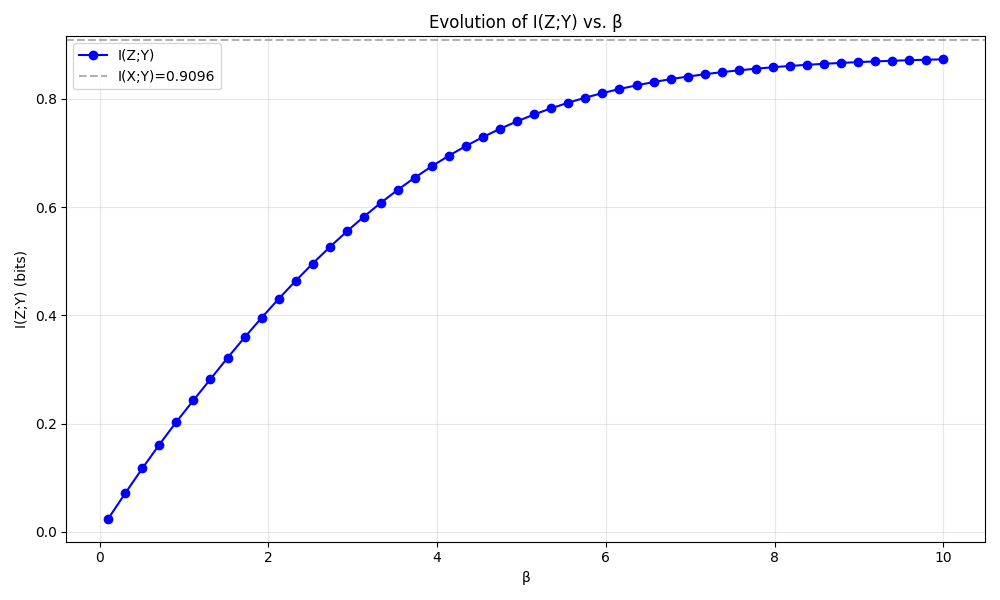}
\caption{Single plot: the saturating curve of $I(Z;Y)$ vs.\ $\beta$. Dotted
gray line for $I(X;Y) \approx 0.9096$. Achieved by $\beta \approx 7$, after
which returns diminish.}
\label{fig:izy_vs_beta_continuation}
\end{figure}

\FloatBarrier
\section{Discussion}
\vspace{0.5em}

\textbf{Smoothness and Stability:}\quad
The primary benefit of my approach is the elimination of unstable jumps in the
IB solution. By convexifying the objective and regularizing entropy, I enforce
smoothness in the encoder as a function of $\beta$. The continuation method
further guarantees that I follow a single, continuous path. This was evident in
both experiments: neither convexified nor entropy-regularized IB exhibited the
dramatic phase transitions that plague the standard IB. From a modeling
perspective, this smoothness means that as I gradually relax compression
(increase $\beta$), the representation $Z$ evolves by splitting clusters
gradually rather than undergoing wholesale reassignments. In the $8 \times 8$
case, for instance, instead of $Z$ jumping from one cluster to two clusters at
a specific $\beta$, I observed fractional membership emerging – effectively,
one cluster softly bifurcated into two, which only became fully distinct
clusters over a range of $\beta$. This is much more akin to human-interpretable
gradual refinement of categories, which could be valuable in applications like
progressive clustering or interactive learning, where one might want to explore
the trade-off continuum.

\vspace{1em}
\textbf{Absence of Instability:}\quad
Standard IB often stalls near phase transitions because the landscape becomes
almost flat along the emerging-cluster direction, forcing Blahut–Arimoto to
crawl or freeze.  In our convexified, entropy-regularized landscape that flat
mode is lifted, so the predictor–corrector scheme keeps a constant step and
converges at the usual speed.  Even if one only wants the optimum at a single
$\beta$, running continuation up to that $\beta$ is safer than random
initialization.  The small entropy term~$\epsilon$ further conditions the
Hessian by preventing $p(z\mid x)$ from hitting exact 0 or~1, eliminating
extreme eigenvalues.  Overall, the solver stays numerically stable and
free of critical slow-downs.
\newpage
\vspace{1em}
\textbf{Comparison of Solvers:}\quad
I compared three solver philosophies: the standard IB solver (BA algorithm per
$\beta$), the convexified solver (my method with $u(t)=t^2$, $\epsilon=0$), and
the entropy-regularized solver (my method with $u(t)=t$, $\epsilon>0$). The
convexified solver yielded the smoothest information curves and strictly convex
$\mathcal{L}$ landscape; its only drawback is that it requires defining a custom
$u$ and the interpretation of $\beta_u$ differs (though I mapped it back to
standard $\beta$ for reporting). The entropy-regularized solver is conceptually
simpler (just adding a regularizer to the IB Lagrangian) and also improved
smoothness, though not as strictly as convexification; it did slightly lag in
achieving full information at a given $\beta$ because it favors some entropy.
However, one can anneal $\epsilon$ down as $\beta$ grows, so that eventually
$\epsilon$ is negligible when approaching the full info regime – this yields
practically the same curve as standard IB but without the jumps. In terms of
implementation, both solvers I presented used the predictor-corrector
continuation. I found that predictor-corrector was crucial for convexified IB
to truly follow the correct branch; if I simply solved (3) independently for
each $\beta$ with random starts, I occasionally got inconsistent branches (as
the objective still doesn't enforce which cluster corresponds to which in a
multi-cluster scenario). The continuation ensures consistency (e.g., it keeps
track of cluster identities across $\beta$). For entropy-regularized IB, one
could in principle run a single optimization at each $\beta$ since the
landscape is smoother, but continuation still offered speed benefits and a
guarantee of monotonic improvement.

\vspace{1em}
\textbf{Multi-path Exploration:}\quad
A fascinating outcome relates to the idea of exploring multiple solutions. In
standard IB, because of non-convexity, multiple local optima for $p(z|x)$ may
exist at a given $\beta$ (especially right after a transition, where one might
find both a solution with $m$ clusters and one with $m+1$ clusters). My
continuation approach inherently chooses one path (the one continuously
connected to the trivial solution). One might ask: do I miss other potentially
better paths? For the IB objective, the answer is generally no – the continuous
path I follow is the global optimum for each $\beta$ by construction, aside
from numerical issues. However, in scenarios with symmetrical structures,
multiple equivalent optima exist (e.g., two ways to assign labels to identical
clusters). My algorithm breaks such symmetry arbitrarily (by slight numerical
perturbations or order of cluster indices) and follows one. If needed, one
could run the solver with different small asymmetric perturbations at $\beta=0$
to obtain other symmetry-related branches (like giving one class a slight bias
to split first). This would yield multiple solution paths, all stable, which
could then be compared in terms of objective (they should be equal if just
relabelings). Therefore, while my goal was a single stable path, the
methodology I developed could be extended to do a mild search over possible
branchings in a controlled manner (as opposed to the standard IB where one has
to guess the number of clusters or restart the algorithm hoping to find
different optima). This is particularly relevant in problems with many symmetric
classes where multiple cluster merge orders are possible. The symbolic
continuation could be guided by domain knowledge (e.g., if one expects certain
splits first). This flexibility is an interesting avenue for future work on
using continuation to map out the entire solution tree of IB, not just the main
branch.

\vspace{1em}
\textbf{Connection to Gaussian and Variational IB:}\quad
Although I focused on discrete IB, the ideas generalize. For continuous or
high-dimensional $X,Y$ (e.g., Gaussian IB or the Variational IB (VIB) for
neural networks \cite{Ref3}), one also faces trade-offs between compression and
prediction. Phase transitions can occur in Gaussian mixture settings or in
training of deep nets (as noted by Shwartz-Ziv \& Tishby \cite{Ref2}, there are phases
where compression kicks in). The convexification concept I propose could be
applied by adjusting the VIB loss to have a convex function of the KL term
(which represents $I(X;Z)$) or adding an entropy term (which in VIB corresponds
to adding noise to the encoder distribution). These could smooth training
dynamics. The implicit ODE approach might also be adapted to track how a neural
network's representation changes with regularization weight – possibly
informing adaptive training schedules. Moreover, the method I propose bears
resemblance to homotopy methods in optimization, where one solves an easier
problem first (here $\beta=0$ trivial solution) and then slowly morphs it into
the harder target problem, staying in the solution manifold. This connection to
continuation methods suggests a link with information geometry: the path
$q(\beta)$ I trace lies on the manifold of probability distributions, and I
ensure it changes smoothly. Information-geometric quantities like Fisher
information might relate to my Hessian-based detection (indeed, Wu et al. \cite{Ref4}
tie the IB phase transition condition to the Fisher information matrix of a
parametrized model). The entropy regularizer I use can be seen as moving on a
geodesic of the entropy (which is concave) to avoid sharp turns. These
interpretations reinforce that the algorithm I developed is implicitly
performing a geometrically natural path tracking on the distribution space.

\vspace{1em}
\textbf{Limitations:}\quad
One limitation is that the convexified objective introduces a hyperparameter
choice – the form of $u(t)$. I used $t^2$ as a simple convex function. In
theory any convex increasing function works \cite{Ref6}; in practice, the degree of
convexity will affect how evenly the IB curve is parameterized by $\beta$. Too
strong convexity (e.g., $u(t)=e^t$) might overly penalize $I(X;Z)$ and keep it
too low until larger $\beta$. Too weak (nearly linear) might not avoid the
degeneracy. I found $t^2$ to be a balanced choice. The entropy regularizer
$\epsilon$ is another parameter – if set too high, the encoder remains very
randomized and one might need higher $\beta$ to differentiate classes
(essentially it biases towards more compression than needed). I used a small
$\epsilon$ ($0.01$–$0.1$) which was enough to smooth things without
significantly altering the trade-off. In principle, one can start with higher
$\epsilon$ at low $\beta$ (to enforce stability when clusters are merging) and
anneal it down to $0$ as $\beta$ grows large (when clusters are well-separated
and stable). I did not implement an automated schedule, but this could combine
the best of both worlds: maximum smoothness early on, and asymptotically exact
IB objective later. Another consideration is computational: while my experiments
were small scale, the Hessian computation may be expensive for very large $|X|$
or continuous spaces. Approximations or quasi-Newton methods could be used.
However, note that if one is already computing the IB solution for many $\beta$
(as often done to plot the IB curve \cite{Ref1}), the method I propose is likely more
efficient overall, since it reuses information (the previous solution and
Hessian) to warm-start the next.
\newpage
\FloatBarrier
\section{Conclusion and Future Work}
\vspace{0.5em}

I presented a novel approach to optimizing the Information Bottleneck objective
that addresses the long-standing issue of solution instability and abrupt phase
transitions. By introducing a convexified IB Lagrangian (e.g., using a squared
mutual information term) and adding entropy regularization, I convexify and
smooth the objective landscape, ensuring a unique and stable optimum for each
trade-off level. Building on these modifications, I developed a continuation
algorithm that symbolically tracks the optimal encoder as the $\beta$ parameter
varies, using implicit differentiation to predict changes and corrector steps
to stay on the optimum path. This method effectively eliminates the
discontinuous jumps (pitchfork bifurcations) that occur in standard IB,
yielding a smooth trajectory in the information plane.
\vspace{0.5em}
My experiments on discrete datasets demonstrated clear benefits: the stable IB
solver produced continuous information curves and gradual cluster evolution, in
contrast to the piecewise constant solutions of standard IB. The method
provides more reliable intermediate representations, which can be crucial in
applications requiring incremental learning or fine-tuning of the
compression–prediction trade-off. Theoretically, my work bridges ideas from
bifurcation theory and information theory, showing that one can steer around a
phase transition by altering the objective and following the implicit path. In
doing so, I affirmed recent theoretical insights that convex surrogates to IB
can recover the entire IB curve \cite{Ref6} and that entropy regularizers can prevent
unstable hard clusterings (as was intuitively known from deterministic
annealing practices \cite{Ref9}).

\vspace{0.5em}
There are several avenues for future work. First, extending these ideas to the
Gaussian IB and Variational IB (VIB) frameworks is a natural next step. In the
Gaussian case, one could replace $I(X;Z)$ (which is quadratic in covariance)
with a higher-order function to avoid degeneracies when $Y$ is deterministically
related to $X$. In deep neural network training with a VIB loss, implementing
an adaptive schedule for the $\beta$ coefficient and perhaps an entropy term
(via noise injection) could stabilize training – this aligns with observations
of two-phase training dynamics \cite{Ref2}, and the method I propose might enforce a
more gradual transition between those phases. Second, exploring the use of
higher-order continuation (arc-length parameterization) could allow the solver
to automatically adjust $\beta$ steps to maintain a roughly constant change in
$q$, which would be useful if the curve has nonlinear parameterization. Third,
my focus was on avoiding transitions, but one could intentionally allow
controlled transitions to occur and examine multi-branch solutions. By relaxing
convexity slightly, one might capture scenarios where two distinct
representations are both viable (a form of model multiplicity). Understanding
and characterizing such multiple IB solutions could deepen the connection to
information geometry and phase transitions (e.g., mapping out the entire
bifurcation diagram as $\beta$ and $\epsilon$ vary).

\vspace{0.5em}
Finally, applying the stable IB optimization to real-world problems will be an
exciting direction. For instance, in feature selection or extraction problems
for sensitive systems (like healthcare or finance), one could use the method I
propose to vary the compression level and obtain a suite of models from highly
compressed to less compressed, without worrying that some models are unreliable
due to training instability. Each model would be a point on a smooth curve, so
a practitioner could pick the sweet spot knowing that a slight change in
$\beta$ won't lead to a drastically different model. Overall, by combining
symbolic continuation with convex optimization techniques, I have shown that
the IB principle can be made stable and practical across its entire operating
range, opening doors for more widespread and reliable use of
information-theoretic learning objectives.
\newpage
\FloatBarrier 

\end{document}